# Coronary Artery Semantic Labeling using Edge Attention Graph Matching Network

Chen Zhao[1], Zhihui Xu[2], Guang-Uei Hung[3], Weihua Zhou[1,4*]

1. Department of Applied Computing, Michigan Technological University, Houghton, MI, USA

2. Department of Cardiology, The First Affiliated Hospital of Nanjing Medical University, Nanjing, China

3. Department of Nuclear Medicine, Chang Bing Show Chwan Memorial Hospital, Changhua, Taiwan

4. Center for Biocomputing and Digital Health, Institute of Computing and Cyber-systems, and Health Research Institute, Michigan Technological University, Houghton, MI, USA

Corresponding author:

Weihua Zhou, PhD, Tel: +1 906-487-2666

E-mail address: whzhou@mtu.edu

Mailing address: 1400 Townsend Dr, Houghton, MI 49931

**Abstract**

Coronary artery disease (CAD) is one of the primary causes leading deaths worldwide. The presence of atherosclerotic lesions in coronary arteries is the underlying pathophysiological basis of CAD, and accurate extraction of individual arterial branches using invasive coronary angiography (ICA) is crucial for stenosis detection and CAD diagnosis. However, deep-learning-based models face challenges in generating semantic segmentation for coronary arteries due to the morphological similarity among different types of arteries. To address this challenge, we propose an innovative approach called the Edge Attention Graph Matching Network (EAGMN) for coronary artery semantic labeling. Inspired by the learning process of interventional cardiologists in interpreting ICA images, our model compares arterial branches between two individual graphs generated from different ICAs. We begin with extracting individual graphs based on the vascular tree obtained from the ICA. Each node in the individual graph represents an arterial segment, and the EAGMN aims to learn the similarity between nodes from the two individual graphs. By converting the coronary artery semantic segmentation task into a graph node similarity comparison task, identifying the node-to-node correspondence would assign semantic labels for each arterial branch. More specifically, The EAGMN utilizes the association graph constructed from the two individual graphs as input. A graph attention module is employed for feature embedding and aggregation, while a decoder generates the linear assignment for node-to-node semantic mapping. Based on the learned node-to-node relationships, unlabeled coronary arterial segments are classified using the labeled coronary arterial segments, thereby achieving semantic labeling. A dataset with 263 labeled ICAs is used to train and validate the EAGMN. Experimental results indicate the EAGMN achieved a weighted accuracy of 0.8653, a weighted precision of 0.8656, a weighted recall of 0.8653 and a weighted F1-score of 0.8643. Furthermore, we employ ZORRO to provide interpretability and explainability of the graph matching for artery semantic labeling. These findings highlight the potential of the EAGMN for accurate and efficient coronary artery semantic labeling using ICAs. By leveraging the inherent characteristics of ICAs and incorporating graph matching techniques, our proposed model provides a promising solution for improving CAD diagnosis and treatment.

## 1. Introduction

Coronary artery disease (CAD) has long been the leading cause of death in the United States [1] and is rapidly becoming the top killer in other counties in the world [2,3]. CAD is characterized by the accumulation of cholesterol-rich plaque within the inner layers of the coronary artery wall. This plaque buildup leads to varying degrees of stenosis, or narrowing, in the arteries. As a consequence, blood flow to the myocardium, the heart muscle, becomes restricted, resulting in myocardial ischemia - a condition marked by insufficient oxygen and nutrient supply to the heart. As the blood supply-demand mismatch worsens, symptoms of CAD such as chest pain (angina), shortness of breath, and others may appear. Moreover, instability in the areas of stenosis can lead to acute occlusion of the coronary artery, resulting in a heart attack. Currently, the treatment for significant coronary artery stenosis involves percutaneous coronary intervention (PCI) or coronary artery bypass grafting (CABG) along with aggressive medical therapies [4].

Invasive coronary angiography (ICA) indeed remains the gold standard for diagnosing CAD [5]. It is an established imaging technique that plays a crucial role in both the diagnosis and treatment of heart conditions, particularly CAD. During an ICA procedure, a thin, flexible tube called a catheter is inserted into an artery, usually in the groin or arm, and guiding it through the blood vessels to the heart. Once the catheter is in place, a special dye is injected into the coronary arteries, which makes them visible. These images support cardiologists identifying any blockages or narrowing in the coronary arteries. Depending on the findings, the cardiologists may use the same catheter to perform treatments such as angioplasty or stent placement to open blocked arteries and improve blood flow to the heart. While ICA is highly effective in providing detailed images of the coronary arteries, it is important to acknowledge the limitations associated with subjective visual assessment. Human interpretation of the angiograms can introduce variability and subjectivity into the diagnosis, potentially leading to unreliable assessments [6].

The coronary vascular tree is indeed complex and contains two major systems: the left coronary artery (LCA) and the right coronary artery (RCA) systems. The LCA is more clinically relevant, given it provides most of the blood supply to the left ventricle. LCA system and codominance are associated with modestly increased PCI in-hospital mortality in patients with stable CAD [7]. According to the 15-segments model defined by the American Heart Association [8], the LCA system can be further subdivided into three main coronary arteries: the left anterior descending (LAD) artery, the left circumflex (LCX) artery, and the left main artery (LMA), which act as the main blood suppliers of the myocardium. In normal coronary anatomy, the LMA bifurcates into two main branches: LAD and LCX. The LAD artery supplies blood to the front and part of the left side of the myocardium muscle while the LCX courses around the left side of the myocardium and supplies blood to the left atrium and part of the left ventricle [9]. The length of LAD varies between 10 to 13 cm and gives rise to the diagonal branches (D), which further contribute to the blood supply of the myocardium. Similarly, the length of the LCX artery varies from 5 to 8 cm and gives rise to obtuse marginal (OM) branches, extending its reach to specific areas of the left ventricle [10].

However, it is challenging to identify individual coronary artery using ICAs because of the morphological similarity among different segments [11] and loss of partial high-frequency detail information due to the projection of the 3D vascular tree into a 2D plane during ICA imaging acquisition [12]. This loss of information limits the ability to precisely discern the intricate structures and boundaries of the coronary arteries, making their identification more challenging. In addition, the coronary arteries not only span over a long distance but also show similar semantic features with each other [12], making it challenging to associate them with the exact branches. This causes the confusion in judgement of whether a coronary segment belongs to main branch or side branch. ICA image understanding is a complex task due to several factors that degrade the visual quality of the images. Furthermore, ICA images suffer from various factors that degrade their visual quality. The contrast degradation occurs as the contrast dye dissipates, leading to reduced visibility of the coronary arteries. Additionally, spatial blurring and overlapping caused by non-vessel tissues and structures further obscure the vessel boundaries and hinder accurate interpretation.

The limitations of existing coronary artery semantic labeling methods that rely solely on position and imaging features are evident when processing complex coronary vasculature from different view angles of ICA images [13]. The morphological similarity among different types of arteries poses a challenge for pixel-intensity-based models to accurately discern each arterial segment and generate meaningful semantic segmentation. To address these challenges, we propose an innovative approach that leverages the concept of graphs to represent the coronary arteries and their connections. By converting the arteries into graph structures, we can incorporate non-traditional features, such as node degrees and graph-theory distances, in addition to the pixel-derived features, to enhance the accuracy of semantic segmentation. A key aspect of our method is the use of graph matching techniques, which aim to find the semantic correspondence between arteries in the labeled graphs. By comparing the graph structures and identifying similar patterns, we can establish meaningful associations between arterial segments and assign appropriate semantic labels.

In this paper, we propose a novel algorithm to perform coronary artery semantic labeling using ICA images. We propose an edge attention graph matching network (EAGMN) to build the semantic correspondence between coronary arterial segments from ICAs. The problem of semantic segmentation is converted into a problem that classifies the type of an unlabeled arterial segment by searching for a labelled arterial segment with the maximal similarity in a database of arterial graphs generated from a large number of ICAs. In detail, the individual graph of the coronary artery is generated according to coronary artery binary segmentation result. The node in the individual graph represents a segment of the coronary artery and the edge indicates the connectivity between arterial segments according to the physical connection of the vascular tree. Then an association graph is constructed from two individual graphs, where each vertex is built from two nodes in the individual graph, representing the node-to-node correspondence of two arterial segments from two individual graphs. Thus, the coronary artery semantic labeling task is converted into vertex classification task using the generated association graph. EAGMN incorporates an encoder module responsible for embedding vertex and edge features using a graph attention convolution network. This module extracts meaningful representations from the association graph by considering the interactions and relationships between the vertices and edges. Additionally, the EAGMN employs a decoder module that facilitates the readout of the feature representations generated by the encoder. By examining the positive vertices in the association graph, which indicate matched nodes from the individual graphs, the EAGMN accomplishes the semantic labeling task by assigning appropriate labels to the coronary arterial segments. The workflow of the proposed EAGMN for coronary artery semantic labeling is shown in Figure 1.

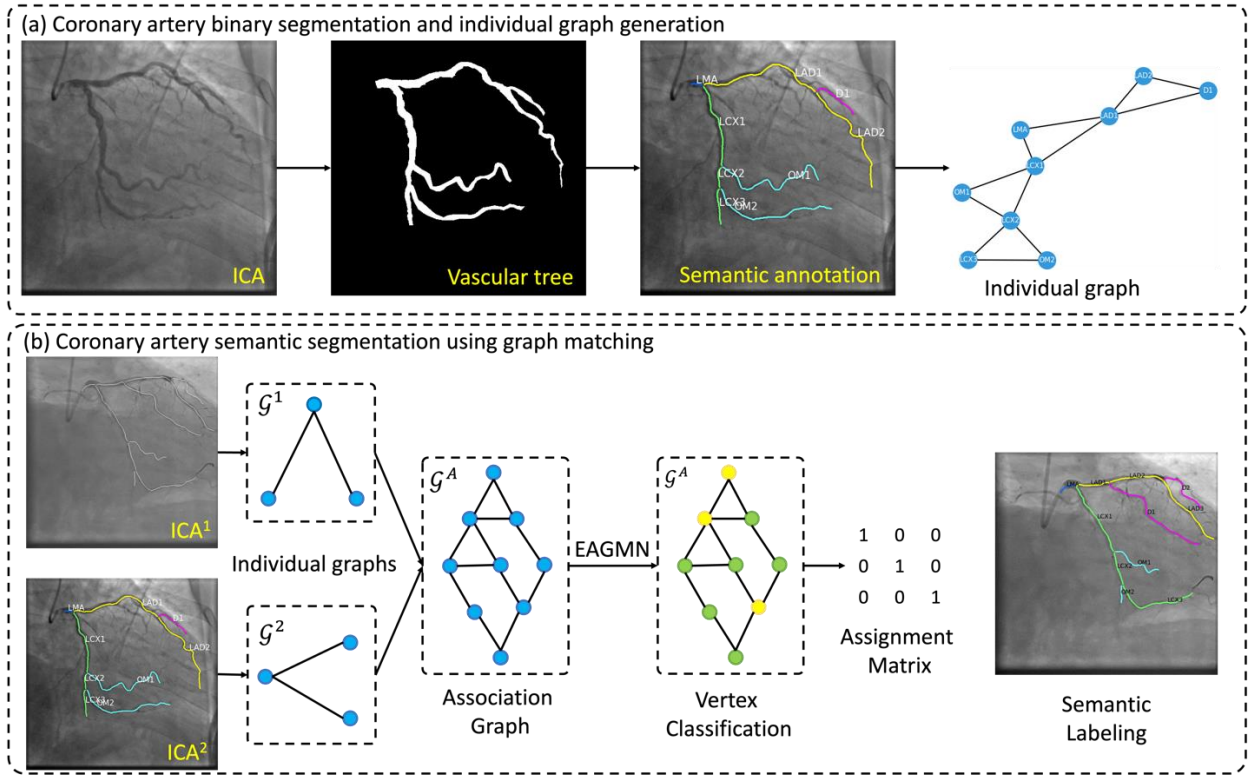

**Figure 1**. Workflow of the proposed EAGMN for coronary artery semantic labeling using EAGMN. (a) Coronary artery binary segmentation by our previous work, feature pyramid U-Net++ [14], and individual graph generation. (b) Coronary artery semantic labeling using EAGMN by comparing the artery-to-artery correspondence between the unlabeled artery in $\mathcal{G}^1$ and labeled arteries in $\mathcal{G}^2$. The positive vertex in yellow and '1' in assignment matrix represent two arterial branches are matched and the type of the unlabeled artery is assigned.

The highlights of this paper are shown below:

1) The utilization of graph matching for coronary artery semantic labeling: This paper introduces a novel approach that employs graph matching techniques to establish the node-to-node correspondence between labeled and unlabeled coronary arterial segments. This enables accurate and reliable semantic labeling of the coronary artery.

2) The EAGMN utilizes an edge attention mechanism to dynamically aggregate features of adjacency edges, which in turn update vertex features for vertex classification using the association graph.

3) We demonstrate the robustness of the proposed model by conducting experiments using corrupted ICAs, which simulates challenging real-world scenarios.

4) We employ ZORRO to provide interpretability and explainability of the graph matching model, shedding light on the decision-making process.

This paper is organized as follows. Section 1 introduces the background, challenges of coronary artery semantic labeling and the highlights of the paper. Section 2 reviews existing algorithms on coronary artery semantic labeling and state-of-the-art graph matching algorithms. In Section 3, the proposed EAGMN is described in detail. The enrolled subjects, implementation details, experimental results, and discussion are presented in Section 4. Section 5 illustrates the conclusion.

**2. Related work**

## 2.1. Coronary artery semantic labeling

Coronary artery semantic labeling can be categorized into pixel-to-pixel/voxel-to-voxel based semantic segmentation methods and segment identification based semantic labeling methods. Pixel-to-pixel based methods focus on achieving dense pixel-level or voxel-level classification by mapping each individual pixel or voxel to its corresponding semantic category. On the other hand, segment identification based methods take a different approach by classifying the entire arterial segment as a whole into a specific semantic category.

Pixel-to-pixel based methods are straightforward because the algorithm assigns unique labels to each pixel in the arterial images. Jun et al. proposed a T-Net for main artery segmentation using ICAs and T-Net achieved a Dice coefficient similarity of 0.8377 using 4700 ICAs for LAD, LCX and RCA segmentation [15]. Xian et al. proposed a U-Net based residual attention network, which integrates attention mechanism and stacks multiple attention modules for main arteries segmentation. The model achieved an F1 score of 0.921 using an dataset with 3,200 ICAs [6]. Zhang et al. proposed the progressive perception learning framework to capture the long-distance semantic relationship, enhance the foreground and suppress the background pixels, and highlight the boundary details for main artery semantic segmentation [12]. The model was validated using a dataset containing 1085 subjects with ICAs and a Dice coefficient similarity of 0.9585 was achieved. Although the aforementioned studies have demonstrated impressive performance in segmenting the main arteries, they rely on separate deep learning models for extracting each type of artery and they cannot label side branches, such as D and OM arteries. Focusing solely on the main branches, such as LMA, LAD, and LCX, may not provide adequate support for downstream CAD analysis.

Segment identification based semantic labeling methods usually contain a vascular tree binary segmentation step and a segment classification step. Cao et al. proposed an automatic coronary artery semantic labeling algorithm based on blood flow and logical rules using coronary computed tomography angiography (CCTA) images [16]. Wu et al. extracted the arterial features according to the spatial locations and directions and organized the arterial segments as tree-structured sequential data. Then, a bi-directional tree-structured long short-term memory network (LSTM) was employed to classify each segment to perform coronary artery semantic labeling [17]. Yang et al. integrated the imaging features extracted by convolutional LSTM and position features as the feature embedding and employed a partial-residual graph convolution network (GCN) for coronary artery semantic labeling [18]. Existing coronary artery semantic labeling methods have demonstrated remarkable performance when applied to 3D CCTA data. However, their direct applicability to 2D ICA images is limited due to the loss of partial high-frequency detail information resulting from the projection of the 3D vascular tree into a 2D plane. Consequently, the labeling and segmentation of coronary arteries in 2D ICA images may suffer from inaccuracies. Our previous work performed coronary artery semantic labeling by learning the semantic correspondence of arterial branches from different individual graphs using graph matching [13]. The information from neighbors is uniformly averaged during the graph matching. However, different types of arteries contribute unequally which induced a lower accuracy. It is crucial to develop methods specifically designed for 2D ICA images to enhance the accuracy and reliability of coronary artery disease diagnosis and treatment.

## 2.2. Graph matching

Graph-matching is a combinatorial optimization problem based on graph structure, which is NP-hard or practically intractable for large-scale settings. The graph matching aims to establish a meaningful node-to-node and edge-to-edge correspondence between different graphs, which is often formulated as a graph edit problem [19], subgraph isomorphism problem and quadradic assignment problem [20]. Traditional graph matching algorithms primarily emphasize combinatorial matching techniques, which involve comparing and matching the structural components of graphs. On the other hand, learning-based methods take a different approach by incorporating feature extraction and affinity learning. These methods are particularly advantageous when dealing with large-scale and high-dimensional data, as they enable the identification of meaningful patterns and relationships within graphs [21]. With the advert development of graph neural network (GNN), it has been embedded into the learning-based graph matching problem because GNN models the structured information and helps transform the graph matching problem into a linear assignment task [22]. Nowak et al. employed GNN to extract structured features of nodes and employed Sinkhorn network as a differentiable layer to solve the linear assignment problem [23]. Wang et al. employed the GNN for both intra-graph node embedding and cross-graph node embedding iteratively for graph matching [24]. Later, Wang et al. directly employed the association graph induced affinity matrix for embedding learning and the Koopmans-Beckmann's QAP was adopted for affinity learning [25]. While existing graph matching algorithms have demonstrated promising performance in various domains, such as nature images, their application to medical images with topological features is still relatively limited and understudied. Given the significance of topological features in medical image analysis, there is a growing need for dedicated research efforts to develop graph matching algorithms specifically tailored to medical imaging. These algorithms should consider the complex interplay between structural elements and leverage the intrinsic knowledge of anatomical connectivity to achieve more accurate and meaningful matching results.

## 3. Methodology

The approach outlined in this study centers on segment identification-based coronary artery semantic labeling. Our proposed EAGMN aims to establish the node-to-node correspondence between two individual graphs, taking into account the edge information. In this framework, each node represents a coronary arterial segment. Consequently, the task of coronary artery semantic labeling is transformed into the problem of finding one-to-one or one-to-zero mappings for arterial segments from two distinct vascular trees. Our method consists of two main steps: vascular tree binary segmentation and segment classification.

### 3.1. Vascular tree extraction and graph generation

Our previous work, Feature Pyramid U-Net++ (FP-U-Net++) [14], is used to extract the vascular tree. To analyze the arterial anatomy and perform the semantic labeling, it is essential to extract the arterial centerline from the extracted vascular tree. The arterial graph generation includes the centerline extraction and arterial segment separation. Centerline extraction is the process of removing the redundant foreground pixels using binary image while preserving the connectivity and the topology of the vascular tree. We adopt the hit-and-mass transformation algorithm [26], which is based on morphological thinning, to extract the arterial centerline. The edge linking algorithm is employed to identify the end points and bifurcation points within the arterial segments. The end points denote the termination of an arterial segment, while the bifurcation points represent the junctions where the arterial segment branches out into sub-branches [27]. Based on the detected end points and bifurcation points, the arterial centerline is divided into distinct centerline segments. Each centerline segment is then associated with an arterial segment, and the semantic labeling involves assigning a label to each arterial segment. An example of the individual graph generation pipeline is shown in Figure 2.

The individual coronary arterial graph is constructed based on the extracted arterial segments, centerline segments, end points, and bifurcation points. The graph's connectivity, represented by its edges, is determined by the connections between arterial segments. Each node in the arterial graph is formed by the endpoints and the bifurcation points, and the semantic labels should be assigned to each edge. However, in the context of the graph matching neural network, the focus is on establishing the correspondence between nodes. Therefore, in practical terms, we interchange the concepts of nodes and edges in the individual graph. In this adjusted perspective, each node represents an arterial segment, and each edge represents an endpoint or a bifurcation point in the arterial centerline, which preserves the connectivity of the arterial tree.

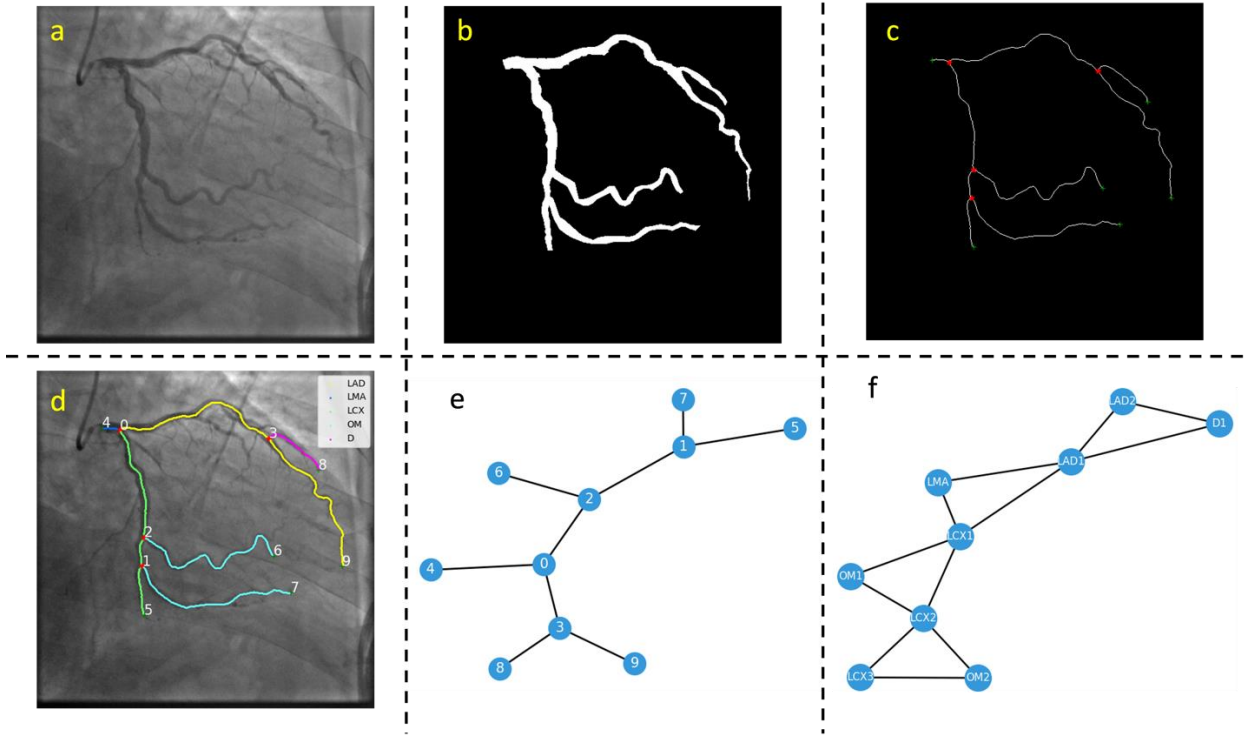

**Figure 2**. Pipeline of the individual graph generation. (a) Original selected ICA frame; (b) Vascular tree generated by FP-U-Net++; (c) Centerline extraction and key points detection; (d) Artery segments with semantic labels; (e) Connectivity of the individual graph; (f) Switched individual graph that each node represents an arterial segment with semantic label and each edge represents the connectivity of the arteries.

Formally, the individual graph represented by an attributed undirected graph as $\mathcal{G} = (\mathbb{V}, \mathbb{E}, \mathcal{V}, \mathcal{E})$, where

- $\mathbb{V} = \{V_i\}$ s.t. $i \in [1, \cdots, n]$ represents node set, and $|\mathbb{V}| = n$.
- $\mathbb{E} = \{E_{ij}\}$ s.t. $i, j \in [1, \cdots, n]$ indicates the set of edges and $|\mathbb{E}| = n_e$.
- $\mathcal{V} = \{v_i\}$ s.t. $i \in [1, \cdots, n]$ indicates the attribute vectors associated with each node.
- $\mathcal{E} = \{e_{ij}\}$ s.t. $i, j \in [1, \cdots, n]$ indicates the attribute vectors associated with each edge.

### 3.2. Graph matching for coronary artery semantic labeling

Graph matching aims to establish a meaningful node-to-node correspondence between individual graphs. Given two undirected attributed individual graphs $\mathcal{G}^1 = (\mathbb{V}^1, \mathbb{E}^1, \mathcal{V}^1, \mathcal{E}^1)$ and $\mathcal{G}^2 = (\mathbb{V}^2, \mathbb{E}^2, \mathcal{V}^2, \mathcal{E}^2)$, where $|\mathbb{V}^1| = n_1$ and $|\mathbb{V}^2| = n_2$, and $|\mathbb{E}^1| = n_{e1}$ and $|\mathbb{E}^2| = n_{e2}$. We aim to find the node correspondence between them by considering the node-to-node affinities and edge-to-edge affinities. Without loss of the generality, we assume $n_1 \leq n_2$. The node-to-node correspondence can be represented by an assignment matrix $M \in \{0,1\}^{n_1 \times n_2}$ that is defined in Eq. 1.

$$M_{ij} = \begin{cases} 1 & if\ V_i\ matches\ V_j \\ 0 & otherwise \end{cases} \quad (1)$$

Inspired by the success of applying association graphs in combinatorial optimization [25,28], we transform the graph matching problem between two individual graphs $\mathcal{G}^1$ and $\mathcal{G}^2$ into a vertex binary classification problem using an attributed undirected association graph $\mathcal{G}^A = (\mathbb{V}^A, \mathbb{E}^A, \mathcal{V}^A, \mathcal{E}^A)$. $\mathcal{G}^A$ contains all candidate correspondence between nodes from two individual graphs. Each vertex of $\mathcal{G}^A$ is built by the nodes of the individual graphs that $\mathbb{V}^A = \{(V_1^1, V_1^2), (V_1^1, V_2^2), \cdots, (V_{n_1}^1, V_{n_2}^2)\}$ and $|\mathbb{V}^A| = n_1 \times n_2$. And each edge in $\mathcal{G}^A$ is constructed by the connectivity of the edges in individual graphs. In detail, if edge $E_{ij}$ ($i, j \in [1, \cdots, n_1]$) from $\mathcal{G}^1$ and $E_{ab}$ ($a, b \in [1, \cdots, n_2]$) from $\mathcal{G}^2$ exist, then edge $E_{ia,jb}$ is constructed in $\mathcal{G}^A$. As such, $|\mathbb{E}^A| = 2 \times n_{e1} \times n_{e2}$. If the node $V_i \in \mathcal{G}^1$ and node $V_a \in \mathcal{G}^2$ have the same semantic labels, then the vertex $V_{ia} \in \mathcal{G}^A$ is a positive vertex, indicating that these two nodes are matched and these two arterial segments are matched; and verse visa. Our association-graph based edge attention graph matching network for coronary artery semantic labeling contains 6 modules, as shown in Figure 3.

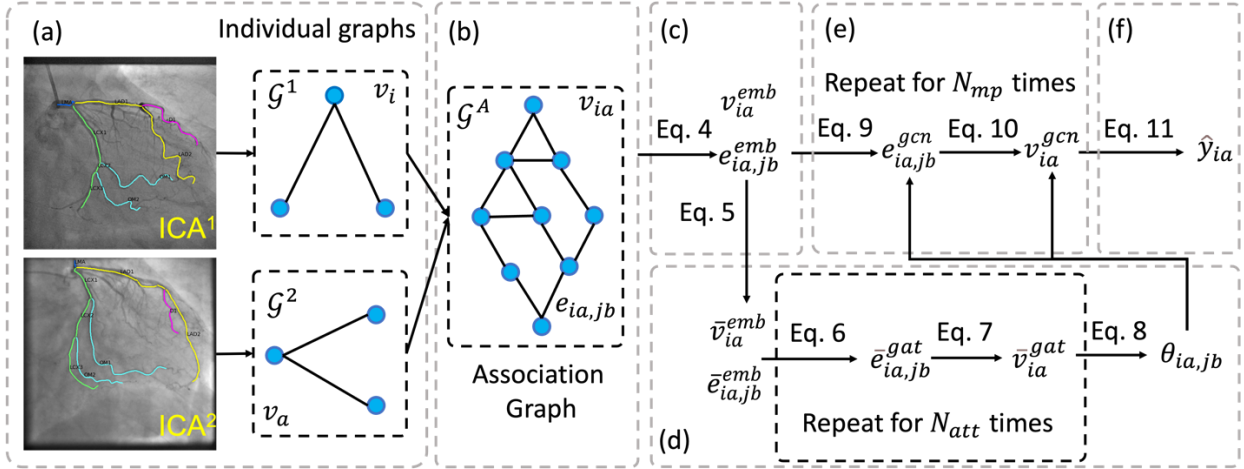

**Figure 3**. Architecture of EAGMN for coronary artery semantic labeling. (a) Individual graph generation and feature extraction; (b) Association graph generation and feature extraction; (c) Vertex and edge feature embedding; (d) Edge attention score calculation using GCN; (e) Feature representation learning according to edge attention score derived in (d); (f) Vertex classification according to the learned feature representations.

*1) Feature extraction in individual graphs.* According to the extracted coronary artery segment, segment centerline, endpoints, bifurcation points and topology of the vascular graph, we extract the imaging features, arterial position features, and topology features for each node in individual graph. For the imaging features, we extract the texture features [11,29] according to masked ICA images for the artery using PyRadiomics [30], including first-ordered features, shape-based features, and gray-level based features. To measure the absolute position of the arterial segment relative to the entire vascular tree, we design 20 hand-crafted position features. These features capture various spatial attributes, such as the distance from the segment to the root of the tree, and the relative position within the tree hierarchy [11]. For the topology feature, we use the degree of the two endpoints of the artery segment as the features. The detailed feature information is shown in Table S1. In total, 121 features are extracted. Without loss of the generality, the dimension of node feature is denoted as $d$ and $v_i \in \mathbb{R}^d$.

*2) Feature extraction in association graph.* The vertex is generated according to the nodes from the individual graphs. Then, the vertex feature $v_{ia}$ is generated by concatenating features from features of node $V_i \in \mathcal{G}^1$ and features of node $V_a \in \mathcal{G}^2$, as defined in Eq. 2.

$$v_{ia} = [v_i, v_a] \in \mathbb{R}^{2d} \text{ s.t. } i \in [1, \cdots, n_1] \text{ and } j \in [1, \cdots, n_2] \tag{2}$$

where $[\cdot]$ denotes feature concatenation. In the same manner, the feature of edge $E_{ia,jb}$ in $\mathcal{G}^A$ is constructed by concatenating the features of edge $E_{ij}$ in $\mathcal{G}^1$ and the features of edge $E_{ab}$ in $\mathcal{G}^2$, as defined in Eq. 3.

$$e_{ia,jb} = [e_{ij}^1, e_{ab}^2] \in \mathbb{R}^{4d} \text{ s.t. } i, j \in [1, \cdots, n_1] \text{ and } a, b \in [1, \cdots, n_2] \tag{3}$$

where $e_{ij}^g = [v_i, v_j]$ s.t. $g \in [1,2]$.

3) Feature embedding. In the feature embedding stage, we aim to transform the input features into a lower-dimensional representation which captures the essential information for the task of coronary artery semantic labeling. This process involves mapping the input features into a feature space where relationships and patterns relevant to the task can be more easily learned and utilized by the model. Using the constructed attribute association graph $\mathcal{G}^A$, two multi-layer perceptron (MLP) based encoders are employed to perform vertex and edge feature embedding, as defined in Eq. 4.

$$\begin{aligned} v_{ia}^{emb} &= f_v^{emb}(v_{ia}) \\ e_{ia,jb}^{emb} &= f_e^{emb}(e_{ia,jb}) \end{aligned} \tag{4}$$

where $f_v^{emb}$ and $f_e^{emb}$ are MLPs with layer-wise instance normalization [31].

*4) Attention score calculation for dynamic feature aggregation.* GCNs are a type of neural networks specifically designed to operate on graph-structured data. The key idea of GCN is to perform convolution on graphs by aggregating information from neighborhood nodes and edges [32]. One of the popular GCNs is graph sample and aggregated (GraphSAGE) GCN [33], which aggregates information from a node's local neighborhood using various aggregation functions, such as mean, max or concatenation. The information from neighbors is uniformly averaged because GCN treats the neighbors equally. However, in coronary artery semantic labeling task, different types of arteries contribute unequally to the central artery segment. Thus, we employed graph attention network (GAT) which incorporates attention mechanisms to improve information aggregation using graph-structured data [34,35]. In detail, an edge attention network is employed to calculate the edge attention score dynamically according to the embedded features in Eq. 4. In detail, the two MLP based encoders are employed to further embed the vertex and edge features, as defined in Eq. 5.

$$\begin{aligned} \bar{v}_{ia}^{emb} &= g_v^{emb}(v_{ia}^{emb}) \\ \bar{e}_{ia,jb}^{emb} &= g_e^{emb}(e_{ia,jb}^{emb}) \end{aligned} \tag{5}$$

where $g_v^{emb}$ and $g_e^{emb}$ are MLPs with layer-wise instance normalization. Then, a GCN layer is employed to perform information aggregation. The GCN layer contains an edge convolution layer and a vertex convolution layer. The edge convolution layer aggregates the information of the connected vertices and updates edge features according to the aggregated features and the previous edge features. Formally, the edge convolution layer contains an aggregation function and an update function, as defined in Eq. 6.

$$\begin{aligned} \bar{e}_{ia,jb}^{gat} &= g_e^{gat}([\bar{v}_{ia}^{emb}, \bar{v}_{jb}^{emb}]) \\ \bar{e}_{ia,jb}^{gat} &\leftarrow g_e^u([\bar{e}_{ia,jb}^{gat}, \bar{e}_{ia,jb}^{emb}]) \end{aligned} \tag{6}$$

where $g_e^{gat}$ is an MLP encoder which aggregates features based on the embedded vertex features in Eq. 5 of the connected two vertices of edge $E_{ia,jb}$; and $g_e^u$ is another MLP encoder which updates edge features according to updated features and the edge features defined in Eq. 5. The purpose of this aggregation process is to enhance the representation of the current edge by considering the characteristics and interactions of the connected vertices. This helps to capture more comprehensive and discriminative information about the graph structure and facilitates the subsequent vertex classification task.

The vertex convolution layer aggregates the information of the connected edges and updates its attributes according to the aggregated features and the embedded features in Eq. 5. Formally, node convolution layer contains an information aggregation layer and an update function, as defined in Eq. 7.

$$\bar{v}_{ia}^{gat} = \sum_{\forall V_{jb} \in A(V_{ia})} g_v^{gat}\left(\bar{e}_{ia,jb}^{gat}\right) \tag{7}$$

$$\bar{v}_{ia}^{gat} \leftarrow g_v^u([\bar{v}_{ia}^{gat}, \bar{v}_{ia}^{emb}])$$

where $g_v^{gat}$ is an MLP encoder which aggregates the updated edge embeddings in Eq. 6 and $A(V_{ia})$ represents the set of the connected vertices of vertex $V_{ia}$. The summation indicates a non-parametric function for feature aggregation. $g_v^u$ is another MLP encoder which updates vertex features according to aggregated edge features and node embedding $\bar{v}_{ia}^{emb}$ defined in Eq. 5.

By iteratively applying Eqs. 6 and 7 for $N_{att}$ times, the edge features are updated according to both the local and global edge and vertex features. Then, a readout module is employed to calculate the edge attention score, as defined in Eq. 8.

$$\theta_{ia,jb} = g_e^{out}(\bar{e}_{ia,jb}^{gat}) \tag{8}$$

where the $g_e^{out}$ is an MLP decoder to calculate the attention score of edge $E_{ij,ab} \in \mathcal{G}^A$, which is denoted as $\theta_{ia,jb} \in \mathbb{R}$. The readout module takes the aggregated features from the previous step and applies further transformations to capture the importance or relevance of each edge. The edge attention score plays a crucial role in guiding the subsequent steps of the model. It helps prioritize the edges that contribute the most to the overall graph structure and assists in making accurate and meaningful vertex classification predictions.

*5) Feature representation learning using GCN.* The feature representation learning module employs a GCN and the calculated graph attention score in Eq. 8. Similarly, this GCN also contains an edge convolution layer and a vertex convolution layer. Formally, edge convolution layer and vertex convolution layer for feature representation learning are defined in Eqs. 9 and 10.

$$e_{ia,jb}^{gcn} = f_e^{gcn}([v_{ia}^{emb}, v_{jb}^{emb}])$$
$$e_{ia,jb}^{gcn} \leftarrow f_e^u([\exp(-\theta_{ia,jb}) \cdot e_{ia,jb}^{gcn}, \exp(-\theta_{ia,jb}) \cdot e_{ia,jb}^{emb}]) \tag{9}$$

$$v_{ia}^{gcn} = \sum_{\forall V_{jb} \in A(V_{ia})} f_v^{gcn}\left(\exp(-\theta_{ia,jb}) \cdot e_{ia,jb}^{gcn}\right)$$
$$v_{ia}^{gcn} \leftarrow f_v^u([v_{ia}^{gcn}, v_{ia}^{emb}]) \tag{10}$$

where $f_e^{gcn}$, $f_e^u$, $f_v^{gcn}$ and $f_v^u$ are MLP encoders; $e^{-\theta_{ia,jb}}$ indicates the exponential form of the edge attention score. By iteratively applying Eqs. 9 and 10 for $N_{mp}$ times, the vertex and edge features are updated according to the neighbors and the edge attention score, and $v_{ia}^{gcn}$ indicates the feature representation of $V_{ia} \in \mathcal{G}^A$.

6) Vertex classification. As mentioned before, the graph matching is equivalent to vertex binary classification. Thus, a vertex classification layer is employed, as defined in Eq. 11.

$$\hat{y}_{ia} = f_{clf}(v_{ia}^{gcn}) \tag{11}$$

where $f_{clf}$ is an MLP decoder with ReLU activation functions and $\hat{y}_{ia} \in \mathbb{R}$ represents the predicted probability of the vertex $V_{ia}$.

### 3.3. Training and testing

To train the proposed EAGMN, we need to prepare pairs of individual graphs, and then generate the association graph based on these selected pairs. In our study, we initially manually annotated the coronary arterial segments using ICA images. Subsequently, we employed the method described in Section 3.1 of our paper to generate the labeled individual graphs. According to the coronary artery anatomy defined by the American Heart Association [8], two cardiologists assigned semantic labels to each arterial segment. Then, the node correspondences between arterial segments are automatically identified and the ground truth of assignment matrix $M$ is generated and $y_{ia} = 1$ if two arterial segments are matched.

However, it should be noted that the main arterial branches, such as LAD and LCX, are often divided into multiple segments due to the presence of side branches like D and OM. As a result, the individual graph representation of these main branches consists of multiple nodes, all of which share the same semantic labels. This introduces a unique challenge in the graph matching task, as it transforms into a one-to-many or many-to-many mapping problem. Unlike the simpler one-to-one or one-to-zero mappings, the presence of multiple possible mappings between nodes significantly increases the complexity of the search space [36,37]. To simplify the complexity of the graph matching task, our approach focuses on one-to-one graph matching. We handle the issue of multiple nodes representing the separated main branches by assigning semantic labels with increasing indices along the blood flow. By assigning incremental semantic labels, we ensure that each node in the individual graph representing a main branch segment has a unique label. This allows us to establish a one-to-one correspondence between the nodes in the individual graphs, simplifying the graph matching process. For example, if LCX and OM1 exist, then LCX is separated into two arterial segments with the semantic labels of LCX1 and LCX2. Then, each arterial segment in $\mathcal{G}^1$ is only matched with one exact segment in $\mathcal{G}^2$.

For model training, a batch of individual graph pairs is selected, and the association graphs are used as the input of the EAGMN. In our study, we specifically focus on ICAs obtained from two different view angles: left anterior oblique (LAO) and right anterior oblique (RAO). We acknowledge that the anatomy and visual characteristics of the coronary arteries can differ between these two view angles. To ensure consistency and accuracy in our analysis, we select individual graphs exclusively from ICAs captured using the same view angle. This allows us to maintain a more homogenous dataset for our analysis and ensure reliable results in coronary artery semantic labeling. The mean squared error (MSE) between the predicted vertex classification probability and the ground truth is used to optimize the model, as defined in Eq. 12.

$$L = MSE(M, \widehat{M}) = \sum_{i=1}^{n_1} \sum_{a=1}^{n_2} (M_{ia} - \widehat{M}_{ia})^2 \qquad (12)$$

where $M_{ia} = y_{ia}$ is the ground truth and $\widehat{M}_{ia} = \hat{y}_{ia}$ is the predicted vertex class. An Adam optimizer with the learning rate of 0.0001 was used to optimize the model. The training algorithm is shown in Algorithm 1.

*Algorithm 1.* The training process of our proposed EAGMN.

---
**Input**:
$D_{tr} = \{\mathcal{G}^i = (\mathbb{V}^i, \mathbb{E}^i, \mathcal{V}^i, \mathcal{E}^i)\}_{i=1}^{n_{tr}}$: training set contains $n_{tr}$ individual graphs with labeled arteries
$N_{att}$: number of message-passing steps for graph attention score calculation
$N_{mp}$: number of message-passing steps for graph convolution
$N$: number of the training iterations
**Output**:
$\Phi$: Trained EAGMN
For $iter = 1 \cdots N$ do
    1. Random select two individual graphs $\mathcal{G}^1$ and $\mathcal{G}^2$ from the same view angles.
    2. Construct association graph $\mathcal{G}^A$ using $\mathcal{G}^1$ and $\mathcal{G}^2$ with Eqs. 2 and 3.
    3. Perform feature embedding for $v_{ia}^{emb}$ and $e_{ia,jb}^{emb}$ with Eq. 4.
    4. Embed vertex and edge features $\bar{v}_{ia}^{emb}$ and $\bar{e}_{ia,jb}^{emb}$ with Eq. 5.
    For $u = 1 \cdots N_{att}$ do
        5. Update edge features $\bar{v}_{ia}^{gat}$ using Eq. 6 and vertex features $\bar{v}_{ia}^{gat}$ with Eq. 7.
    6. Calculate edge attention score $\theta_{ia,jb}$ with Eq. 8.
    For $v = 1 \cdots N_{mp}$ do
        7. Update edge features using $e_{ia,jb}^{gcn}$ with Eq. 9 and vertex features $v_{ia}^{gcn}$ with Eq. 10
    8. Perform vertex classification with Eq. 11.
    9. Optimize EAGMN using Eq. 12.

---

During the model testing phase, we employ a voting strategy to assign semantic labels to artery branches in the tested ICAs. Our proposed EAGMN establishes the node-to-node semantic correspondence between the individual graphs. During the testing, $\mathcal{G}^1$ is generated using ICAs in the testing set and $\mathcal{G}^2$ is generated using ICAs in the template set, where template set contains a set of representative ICAs selected by experienced cardiologists. We simulate the learning procedure by graph matching that the cardiologist learns the coronary artery anatomy by comparing the anatomy of the testing case with the reference cases in the template set. By applying the EAGMN to the association graph generated by the tested ICA and template ICA, the artery segments are matched, and the semantic labeling is achieved. Given the complex and diverse structure of coronary artery anatomy, clinical decisions are often made based on multiple ICA frames. Therefore, we perform graph matching between the tested ICA and every ICA in the template set. Each arterial segment of the tested ICA is matched to multiple arterial segments in the template set, and a majority voting strategy is employed to assign the label according to the labels of the matched arterial segments in the template set. This approach helps ensure robustness and accuracy in the final semantic labeling process. The testing algorithm is shown in Algorithm 2.

*Algorithm 2.* The testing process of our proposed EAGMN.

---
**Input**:
$D_{te} = \{\mathcal{G}^i = (\mathbb{V}^i, \mathbb{E}^i, \mathcal{V}^i, \mathcal{E}^i)\}_{i=1}^{n_{te}}$: testing set contains $n_{te}$ individual graphs without labeled arteries.
$D_{tp} = \{\mathcal{G}^j = (\mathbb{V}^j, \mathbb{E}^j, \mathcal{V}^j, \mathcal{E}^j)\}_{j=1}^{n_{tp}}$: template set contains $n_{tp}$ individual graphs with labeled arteries.
$\Phi$: trained AGAMN
**Output**:
Labels for each segment of $n_{te}$ ICA graphs in $D_{te}$
For $i = 1 \cdots n_{te}$ do
    For $i = j \cdots n_{tp}$ do
        If $|\mathbb{V}^i| \leq |\mathbb{V}^j|$ and $\mathcal{G}^i$ and $\mathcal{G}^j$ are from the same view angles, then
            1. Construct association graph $\mathcal{G}^A$ using $\mathcal{G}^i$ and $\mathcal{G}^j$ with Eqs. 2 and 3
            2. Calculate assignment matrix $\widehat{M}^{ij} = \Phi(\mathcal{G}^A)$
        3. Assign labels to $\mathcal{G}^i$ according to the majority voting among set of the $\widehat{M}^{ij}$, s.t. $j \in [1, \cdots n_{tp}]$

---

### 3.4. Evaluation

The semantic labeling problem is converted into a multi-class classification problem among arterial segments. As a classification problem, the weighted accuracy (ACC), weighted precision (PREC), weighted recall (REC), and weighted F1-score (F1) are used to evaluate the model performance. We separate the LAD and LCX branches into sub-segments during the model training. However, we group the sub-segments into their original classes in the evaluation process. The weighted ACC, PREC, REC, and F1 definitions are shown in Eqs. 13 to 16.

$$ACC = \frac{1}{n}\sum_{c=1}^{C} \frac{TP_c + TN_c}{TP_c + TN_c + FN_c + FP_c} \times n_C \tag{13}$$

$$PREC = \frac{1}{n}\sum_{c=1}^{C} \frac{TP_c}{TP_c + FP_c} \times n_C \tag{14}$$

$$REC = \frac{1}{n}\sum_{c=1}^{C} \frac{TN_c}{TN_c + FN_c} \times n_C \tag{15}$$

$$F1 = \frac{1}{n}\sum_{c=1}^{C} \frac{TP_c}{TP_c + \frac{1}{2}(FP_c + FN_c)} \times n_c \tag{16}$$

where $TP_c, TN_c, FP_c$ and $FN_c$ represent the true positive, true negative, false positive, and false negative arterial segments, respectively. $C$ is the total number of classes. $n_C$ is the number of arterial segments in class $C$, and $n$ is the total number of arterial segments.

### 3.5. Model interpretation

Interpreting the decisions made by EAGMN is crucial for building trust and confidence in its predictions, especially in clinical practice. Existing approaches for explaining both node and edge features in GNN focus on gradient-based approaches and perturbation-based methods [38]. Gradient-based methods suffer from saturation problems in that the model output changes minimally with respect to any input change; thus, we adopt a perturbation-based method, ZORRO [39], which compares the output variations with respect to different input perturbations to calculate the node and feature importance for our graph matching network. Given a generated association graph, ZORRO iteratively and recursively adds the important features and nodes according to the fidelity score, where the fidelity score measures the difference between the raw model predictions and the predictions after masking out the important features and nodes [39]. Formally, the fidelity score is defined as

$$F(\mathbb{V}_s, \mathbb{F}_s) = \frac{1}{n_1 n_2} \sum_{i=1}^{n_1} \sum_{a=1}^{n_2} \mathbb{I}(y_{ia}^s = \hat{y}_{ia}) \tag{17}$$

where $\mathbb{I}$ is the indicator function. $\hat{y}_{ia}$ is the original prediction of the model and $y_{ia}^s$ is the prediction using the masked nodes by $\mathbb{V}_s$ and the masked features by $\mathbb{F}_s$. $\mathbb{V}_s$ and $\mathbb{F}_s$ are binary vectors in which 1 indicates the node/feature is selected and vice versa. An explanation for GNN-based model indicates that using the masked features and the masked nodes in graph, the fidelity score measured by the new prediction and the original prediction reached $\tau$, i.e. $\tau \geq F(\mathbb{V}_s, \mathbb{F}_s)$.

*Explaining feature importance.* In section 3.1, we manually designed 121 hand-crafted features. The vertex features are generated by the concatenation of the nodes in the individual graph and identifying the concatenated features for each vertex cannot explain the feature importance for the node in the individual graph. To explain the feature importance, a unified feature mask is applied to the feature of every node in the individual graphs, then the masked features are concatenated as the vertex features using Eq. 2. ZORRO masks out all features at the beginning of the explanation and gradually adding features with the highest fidelity scores until the fidelity score is reached at $\tau$. By applying ZORRO to all individual pairs generated by the testing set and the template set, the frequency of the selected important features indicates the feature importance used for graph matching.

*Explaining node importance.* For the original ZORRO, neighbor nodes are removed first and then gradually added to test the importance of the node by the improvement of the fidelity scores. The aim of explaining node importance is to interpret the importance of different arterial segments for graph matching. During the testing phase, the individual graph $\mathcal{G}^1$ is the tested case and $\mathcal{G}^2$ is the template ICA graph. However, our graph matching algorithm is applied to the association graph, examining the importance of vertex in the association graph cannot reflect the node importance. Thus, we modify the ZORRO to explain the node importance that at the beginning of the explanation, all nodes in $\mathcal{G}^1$ is retained and all nodes in $\mathcal{G}^2$ are removed. ZORRO iteratively adds the node from $\mathcal{G}^2$ with the highest improvement of the fidelity score. According to the improvement of the fidelity score, the importance of the node is obtained. The node importance in our approach reflects the significance of each arterial segment in the template ICAs for both graph matching and semantic labeling. By assigning importance scores to the nodes, we can capture the relevance and influence of each arterial segment in the template set. This allows us to prioritize and weight the contributions of different segments during the graph matching process.

## 4. Experiments and discussions

### 4.1. Dataset

In this study, we manually annotated 204 and 59 ICAs from site 1 at The First Affiliated Hospital of Nanjing Medical University and site 2 at Chang Bing Show Chwan Memorial Hospital, respectively. In total, this retrospective study enrolled 263 ICA images. The detailed description of the image acquisition was illustrated in our previous work [13]. For each patient, a frame that was used for anatomical structure analysis in clinical practice was selected from the view video for semantic labeling. In this study, we only focus on semantic labeling for the main branches of LMA, LAD, and LCX, and the side branches of D and OM.

### 4.2. Implementation details

We implemented our EAGMN using TensorFlow and GraphNets [40]. The vertex feature embedding module $f_v^{emb}$, the edge feature embedding module $f_e^{emb}$, the graph convolution module $f_e^{gcn}, f_e^u, f_v^{gcn}$ and $f_v^u$ were implemented using two MLP layers with layer-wise instance normalization and the number of hidden units were set as 64. For the attention score calculation, the vertex feature embedding module $g_v^{emb}$, the edge feature embedding module $g_e^{emb}$, the graph attention module $g_e^{gat}$, $g_e^u$, $g_v^{gat}$ and $g_v^u$ were also implemented using two MLP layers with the 64 hidden unites. The edge attention score readout module $g_e^{out}$ and the vertex classification layer $f_{clf}$ were implemented using two MLP layers with ReLU activation functions. The number of message-passing steps for graph attention score calculation, $N_{att}$, was set as 3 and the number of message-passing steps for graph convolution, $N_{mp}$, was set as 2.

### 4.3. Coronary artery semantic labeling performance

Our ICA dataset contains 263 annotated ICA graphs, and 79 of them were selected as the template set. The remining 184 ICAs were used for five-fold cross validation, which indicates that each fold 147 ICAs were used as the training set and 37 ICAs were used as the testing set. We trained our EAGMN for $N = 100,000$ steps. The averaged performance on the five-folds were reported. The achieved performance of our EAGMN for coronary artery semantic labeling is shown in Table 1.

**Table 1**. Achieved performance for coronary artery semantic labeling using the proposed EAGMN.

| Artery type | ACC | PREC | REC | F1-score |
|---|---|---|---|---|
| LMA | 1.0000±0.0000 | 0.9841±0.0130 | 1.0000±0.0000 | 0.9919±0.0066 |
| LAD | 0.8905±0.0661 | 0.8692±0.0491 | 0.8905±0.0661 | 0.8795±0.0563 |
| LCX | 0.8831±0.0652 | 0.8444±0.0405 | 0.8831±0.0652 | 0.8629±0.0497 |
| D | 0.8142±0.0714 | 0.8335±0.0821 | 0.8142±0.0714 | 0.8233±0.0741 |
| OM | 0.7597±0.0408 | 0.8472±0.0769 | 0.7597±0.0408 | 0.7999±0.0505 |
| weighted average | 0.8653±0.0488 | 0.8656±0.0499 | 0.8653±0.0488 | 0.8643±0.0485 |

As illustrated in Table 1, our EAGMN achieved an average accuracy of 0.8653, which indicating 86.53% of the arterial segments in the testing set were correctly classified according to the trained EAGMN and the selected ICAs in the template set.

We also compared the EAGMN with three existing segment identification based semantic labeling methods. In addition, we implemented 3 graph matching neural networks and applied them into coronary artery semantic labeling task using the same training and testing algorithms developed in this study.

- SVM [11]: In our previous work, a support vector machine was employed to classify coronary artery segments using the 20 position features and 2 topology features as described in section 3.1.
- BiTreeLSTM [17]: The bidirectional tree LSTM (BiTreeLSTM) was initially applied for coronary artery semantic labeling using CCTA according to the spatial locations and directions of arteries in 3D. We adopted the same network architecture but extracted the spatial locations and directions of coronary arteries in 2D.
- CPR-GCN [18]: CPR-GCN was proposed to perform coronary artery semantic labeling using CCTA with 3D convolution LSTM for arterial imaging feature extraction and GCN for artery semantic labeling. We adopted the same network architecture but extracted the arterial imaging features using 2D convolution LSTM.
- IPCA [24]: Iterative Permutation loss and Cross-graph Affinity (IPCA) for graph matching employed the GNN for both intra-graph node embedding and cross-graph node embedding iteratively for graph matching using individual graphs.
- NGM [25]: Neural Graph Matching (NGM) network employed the association graph induced affinity matrix for embedding learning. In our EAGMN, the connectivity of the association was generated by the connectivity of the individual graphs; however, NGM adopted the Koopmans-Beckmann's QAP for affinity learning to calculate the assignment matrix.
- AGMN [13]: Our previous work adopted the association graph for graph matching without the graph attention module.

For each baseline, we performed five-fold cross validation, and the performance comparisons of the coronary artery semantic labeling are illustrated in Table 2.

**Table 2**. Comparison of coronary artery semantic labeling between the proposed EAGMN and existing methods using our ICA dataset. The achieved highest performance is shown in bold.

| Method | Metric | LMA | LAD | LCX | D | OM | mean |
|---|---|---|---|---|---|---|---|
| ML | ACC | 0.9925±0.0151 | 0.6331±0.0540 | 0.6388±0.0390 | 0.6147±0.0410 | 0.5907±0.0419 | 0.6651±0.0080 |
| BiTreeLSTM | | **1.0000±0.0000** | 0.8845±0.0150 | **0.9871±0.0120** | 0.0000±0.0000 | 0.5981±0.0165 | 0.7492±0.0085 |
| CPR-GCN | | 0.5361±0.2996 | 0.5319±0.1239 | 0.5072±0.1447 | 0.0624±0.0953 | 0.5341±0.3045 | 0.4581±0.0536 |
| IPCA | | 0.9820±0.0220 | 0.8222±0.0699 | 0.7837±0.0634 | 0.7924±0.0857 | 0.7140±0.0676 | 0.8039±0.0537 |
| NGM | | 0.9953±0.0093 | 0.8316±0.0586 | 0.8154±0.0739 | 0.7629±0.0870 | 0.7105±0.1036 | 0.8130±0.0636 |
| AGMN | | 0.9956±0.0089 | 0.8432±0.0306 | 0.8046±0.0452 | 0.7956±0.0412 | 0.7565±0.0825 | 0.8264±0.0302 |
| EAGMN | | **1.0000±0.0000** | **0.8905±0.0661** | 0.8831±0.0652 | **0.8142±0.0714** | **0.7597±0.0408** | **0.8653±0.0488** |
| ML | PREC | 0.9778±0.0071 | 0.6586±0.0174 | 0.6375±0.0378 | 0.5554±0.0101 | 0.6278±0.0261 | 0.6679±0.0081 |
| BiTreeLSTM | | **1.0000±0.0000** | 0.8562±0.0190 | 0.5853±0.0099 | 0.0000±0.0000 | **0.9808±0.0122** | 0.6927±0.0074 |
| CPR-GCN | | 0.6208±0.3240 | 0.5675±0.0540 | 0.3964±0.0570 | 0.2802±0.3727 | 0.3821±0.0139 | 0.4463±0.1075 |
| IPCA | | 0.9779±0.0135 | 0.7865±0.0634 | 0.8008±0.0552 | 0.7233±0.0572 | 0.7932±0.0673 | 0.8046±0.0545 |
| NGM | | 0.9910±0.0110 | 0.8269±0.0706 | 0.7985±0.0669 | 0.7546±0.0688 | 0.7492±0.0866 | 0.8122±0.0644 |
| AGMN | | 0.9911±0.0109 | 0.8476±0.0481 | 0.8256±0.0307 | 0.7536±0.0493 | 0.7613±0.0319 | 0.8276±0.0298 |
| EAGMN | | 0.9841±0.0130 | **0.8692±0.0491** | **0.8444±0.0405** | **0.8335±0.0821** | 0.8472±0.0769 | **0.8656±0.0499** |
| ML | REC | 0.9925±0.0151 | 0.6331±0.0540 | 0.6388±0.0390 | 0.6147±0.0410 | 0.5907±0.0419 | 0.6651±0.0080 |
| BiTreeLSTM | | **1.0000±0.0000** | 0.8845±0.0150 | **0.9871±0.0120** | 0.0000±0.0000 | 0.5981±0.0165 | 0.7492±0.0085 |
| CPR-GCN | | 0.5361±0.2996 | 0.5319±0.1239 | 0.5072±0.1447 | 0.0624±0.0953 | 0.5341±0.3045 | 0.4581±0.0536 |
| IPCA | | 0.9820±0.0220 | 0.8222±0.0699 | 0.7837±0.0634 | 0.7924±0.0857 | 0.7140±0.0676 | 0.8039±0.0537 |
| NGM | | 0.9953±0.0093 | 0.8316±0.0586 | 0.8154±0.0739 | 0.7629±0.0870 | 0.7105±0.1036 | 0.8130±0.0636 |
| AGMN | | 0.9956±0.0089 | 0.8432±0.0306 | 0.8046±0.0452 | 0.7956±0.0412 | 0.7565±0.0825 | 0.8264±0.0302 |
| EAGMN | | **1.0000±0.0000** | **0.8905±0.0661** | 0.8831±0.0652 | **0.8142±0.0714** | **0.7597±0.0408** | **0.8653±0.0488** |
| ML | F1 | 0.9850±0.0076 | 0.6437±0.0213 | 0.6360±0.0087 | 0.5832±0.0234 | 0.6071±0.0183 | 0.6646±0.0077 |
| BiTreeLSTM | | **1.0000±0.0000** | 0.8699±0.0101 | 0.7348±0.0093 | 0.0000±0.0000 | 0.7429±0.0141 | 0.6967±0.0085 |
| CPR-GCN | | 0.5698±0.3026 | 0.5455±0.0899 | 0.4353±0.0632 | 0.0742±0.0957 | 0.3924±0.1660 | 0.4192±0.0661 |
| IPCA | | 0.9798±0.0133 | 0.8038±0.0654 | 0.7921±0.0591 | 0.7559±0.0695 | 0.7510±0.0645 | 0.8033±0.0538 |
| NGM | | 0.9932±0.0092 | 0.8290±0.0637 | 0.8067±0.0699 | 0.7583±0.0760 | 0.7289±0.0946 | 0.8123±0.0640 |
| AGMN | | 0.9933±0.0089 | 0.8452±0.0386 | 0.8143±0.0310 | 0.7736±0.0424 | 0.7569±0.0508 | 0.8262±0.0301 |
| EAGMN | | 0.9919±0.0066 | **0.8795±0.0563** | **0.8629±0.0497** | **0.8233±0.0741** | **0.7999±0.0505** | **0.8643±0.0485** |

According to Table 2, the proposed EAGMN achieved the highest accuracy of coronary artery semantic labeling for 0.8653. However, the BiTreeLSTM baseline achieved stable performance in labeling LMA branches. For BiTreeLSTM and CPR-GCN models, they were initially designed for coronary artery semantic labeling using CCTA datasets. Though they showed impressive performance with an ACC greater than 0.9, they may not be suitable for coronary artery semantic labeling using ICAs. Understanding the relationships between arteries and their spatial orientation requires a three-dimensional perspective; however, coronary artery labeling using 2D images is challenging and these two methods cannot achieve satisfactory performance. In detail, the CPR-GCN achieved a weighted ACC of 0.4581 while the BiTreeLSTM achieved an ACC of 0.7492, which were significantly lower than the graph matching based methods. The results indicate that the complex branching patterns and variations in anatomy make it more challenging to understand coronary artery anatomy, and learning the ICA anatomy using individual graphs is difficult.

For the graph matching based methods, we use IPCA, NGM and AGMN as baselines. IPCA [24] first performs feature embedding using individual graphs, and then performs cross-graph feature embeddings using GCNs. Compared to association graph-based graph matching algorithms, individual graph-based matching methods are computationally efficient and suitable for simpler graphs but may struggle with complex graph structures. IPCA achieved an ACC of 0.8039, which was lower than the association graph-based methods. NGM [25] first embeds features of nodes of the individual graphs to calculate the affinity matrix, and uses the calculated affinity matrix as the adjacency matrix for the association graph. Then, NGM performs the feature embedding according to the node features for vertex classification.

On the contrary, AGMN [13] and our proposed EAGMN concatenate the node features of the individual graphs and perform the node embedding using the concatenated features. The adjacency matrix of the association graph is generated by the connectivity of the individual graphs rather than the affinity matrix used by NGM. AGMN and EAGMN achieved the ACC of 0.8264 and 0.8653, which were higher than that of NGM. Experimental results indicate that the physical connectivity of the coronary artery plays an important role in graph matching, where the physical connectivity refers to the actual anatomical connections between different arterial segments within the coronary artery network. The physical connectivity of the coronary artery provides valuable cues and constraints that guide the matching process. It helps ensure that the matched arterial segments are not only similar in appearance or position but also consistent with the underlying anatomical structure. Compared to AGMN, our proposed EAGMN employs the edge attention mechanisms to aggregate information from the adjacent edges dynamically. By employing edge attention mechanisms, EAGMN can adaptively prioritize and weight the importance of different edges based on their relevance to the graph matching task. This dynamic attention mechanism allows the model to focus on the most informative and discriminative edges while suppressing the influence of less relevant edges. Experimental results indicate that EAGMN improved the ACC by 0.0389 for all types of arteries compared to that of AGMN, which outperformed other baseline models significantly.

### 4.4. Robustness test

In clinical practice, due to the imaging quality, artery overlapping and human subjective annotation, the algorithm proposed in section 3.1 may not generate a perfect coronary arterial graph. To reflect the robustness of the proposed EAGMN and test the clinical flexibility, we performed the robustness test using the corrupted individual graphs with random arterial segment dropping. Note that the graph-based approaches use the topology information and the connected graphs as the input, thus, we only randomly removed the arterial branches with at least one endpoint during the robustness test. In addition, the LMA was retained in all experiments due to the importance of this in-let artery segmentation for the coronary artery anatomy. We set the probability of 5%, 7.5%, 10%, 12.5%, 15%, 17.5% and 20% to randomly remove the arteries, and the averaged performance for the robustness test among baselines and EAGMN of ICAs in testing set is shown in Figure 4.

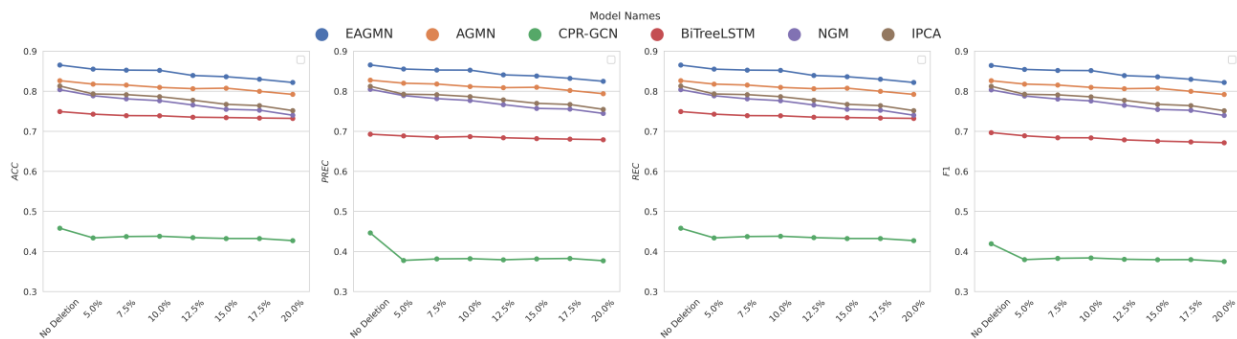

**Figure 4.** Robustness test of the coronary artery semantic labeling among baselines and the proposed EAGMN. The horizontal axis indicates the probability of dropping an artery segment and the vertical axis represents its corresponding performance.

According to Figure 4, the proposed EAGMN was robust since the achieved ACCs were greater than 0.81 during these robust testing experiments, even with 20% missing arterial segments. The BiTreeLSTM and CPR-GCN showed a minor performance dropping, which indicated that these two methods did not fully utilize the graph structure of coronary artery. In comparison to the graph matching-based methods, such as IPCA, NGM, and AGMN, our proposed EAGMN consistently demonstrated superior performance in terms of ACC, PREC, REC, and F1 across various missing probabilities. These results highlight the robustness of the proposed EAGMN in handling missing data scenarios.

### 4.5. Result explanation

We adopted 37 ICAs as the testing set and 70 ICAs as the template set. Each ICA in testing set was used as $\mathcal{G}^1$ to perform graph matching with every ICA as $\mathcal{G}^2$ in template set. Note that number of nodes in $\mathcal{G}^1$ is smaller than $\mathcal{G}^2$, and 1148 individual graph matching pairs were generated. We applied ZORRO to explain each graph matching pair to iteratively and recursively calculate the feature and node importance for graph matching. The threshold of the fidelity score used to measure the new prediction and the original prediction,

$\tau$, was set as 0.8. We ranked the frequency of the selected important features to measure the feature importance, as shown in Figure 5.

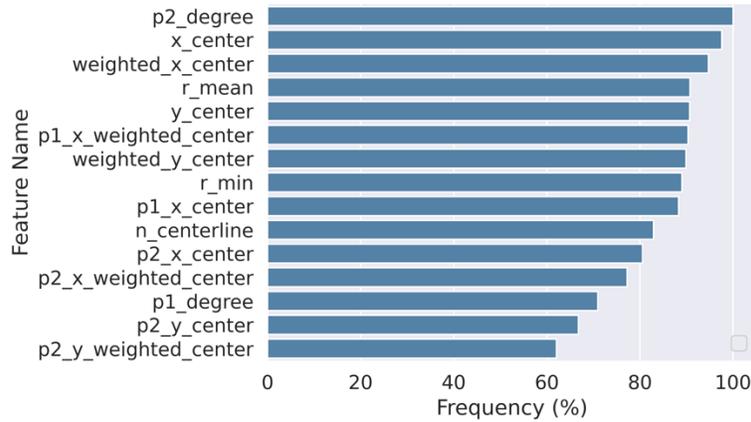

**Figure 5.** Feature importance explained by modified ZORRO. The vertical axis represents the name of the hand-crafted features, and the horizontal axis represents the frequency of the selected features when the explaination was reached.

According to Figure 5. The top 15 features with the most importance are topology-based features, i.e. p2_degree and p1_degree, and position-based features. The p1_degree and p2_degree represent the degree of the endpoint of the left and right end of the arterial segment. The results indicate that the topology is an extremely important factor in arterial identification. The position features capture the absolute and relative position of the arterial segment in relation to the overall vascular tree, which also provide concrete information for artery semantic labeling. This information helps in distinguishing and identifying the individual arterial segments based on their unique spatial locations.

For explaining the node importance, we visualized the graph matching results and the fidelity score improvement when adding the node in the $\mathcal{G}^2$ into the node mask. Example results are shown in Figure 6. In Figure 6, the top four pairs of ICA graphs demonstrate a successful and accurate matching with only minor errors. However, the bottom two pairs do not exhibit a satisfactory level of matching. Upon analyzing the fidelity scores depicted in the pseudo color visualization, we observed that the LMA branch plays a crucial role in the identification of coronary arteries. The successful identification of the LMA branch, along with the first segment of LAD (LAD1) and the first segment of LCX (LCX1), is of paramount importance for the semantic labeling of the entire vascular graph.

In contrast, for the bottom two pairs of graphs, the significance of the LMA branch was diminished compared to the top four pairs, leading to a degradation in performance. The analysis of the node importance further confirms that our EAGMN closely mimics the approach of cardiologists in identifying coronary arteries. It starts by focusing on the main LMA branch and subsequently considers the two most vital subbranches, namely LAD1 and LCX1, in order to achieve accurate semantic labeling. Overall, these findings highlight the importance of the LMA branch and the initial segments of LAD and LCX in the graph matching process. They demonstrate that our EAGMN model effectively emulates the decision-making process followed by cardiologists when identifying and labeling coronary arteries.

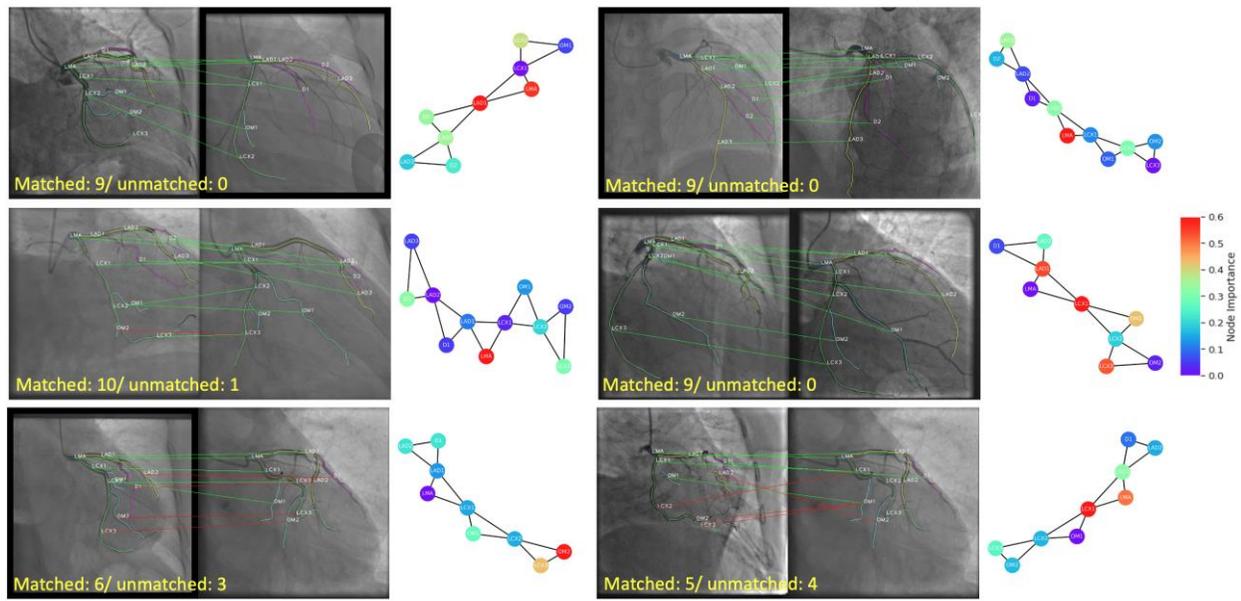

**Figure 6**. Visualization of graph matching results and the improvement of the fidelity score when adding the arterial segment in template graph for graph matching. The green line indicates a correct match, and the red line represents an error.

## 5. Conclusion

In this paper, we present an edge attention graph matching network for coronary artery semantic labeling. By performing the graph matching between individual ICA generated graphs, the relationship between the arterial segments is obtained and the unlabeled coronary arterial segments are labeled by template ICAs. Experimental results showed that our approach is powerful and robust. By employing the ZORRO, we explained the graph matching results and improved the interpretability of coronary artery semantic labeling using our proposed approach.

## Acknowledgement

This research was supported in part by a research seed fund from Michigan Technological University Health Research Institute and an NIH grant (U19AG055373).